\documentclass[11pt]{article}

\usepackage[preprint]{acl}
\usepackage{times}
\usepackage{latexsym}
\usepackage[T1]{fontenc}
\usepackage[utf8]{inputenc}
\usepackage{microtype}
\usepackage{inconsolata}
\usepackage{graphicx}
\usepackage{booktabs}
\usepackage{amsmath}
\usepackage{amssymb}
\usepackage{arydshln}
\setcitestyle{citesep={,},yysep={,}}

\makeatletter
\newcommand{\ps@arxivfirstpage}{%
  \let\@oddhead\@empty
  \let\@evenhead\@empty
  \def\@oddfoot{%
    \makebox[0pt][l]{\normalfont\small\itshape Under review}%
    \hfil{\normalfont\thepage}\hfil
  }%
  \let\@evenfoot\@oddfoot
}
\makeatother

\newcommand{\methodoverviewgraphic}{%
  \includegraphics[width=\textwidth,pagebox=cropbox]{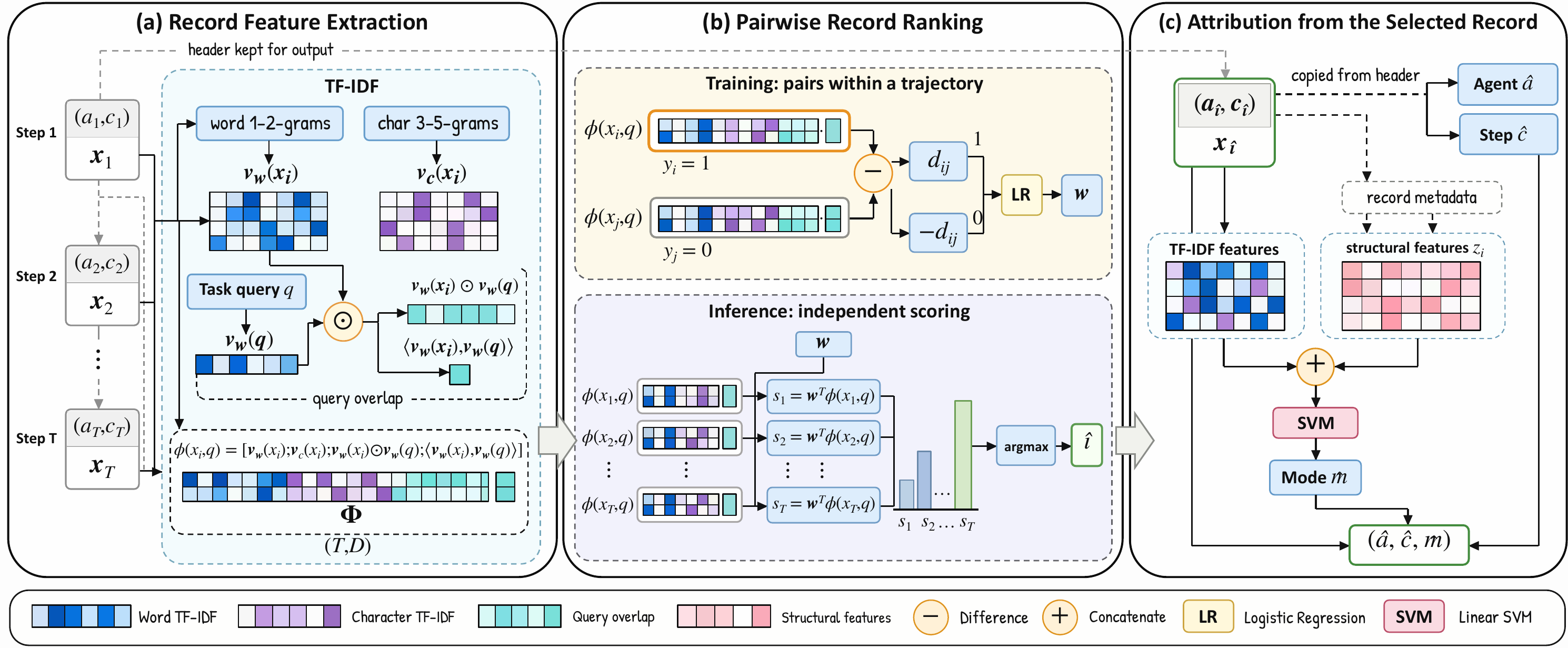}%
}
\newcommand{\traceexamplegraphic}{%
  \includegraphics[width=\columnwidth,pagebox=cropbox]{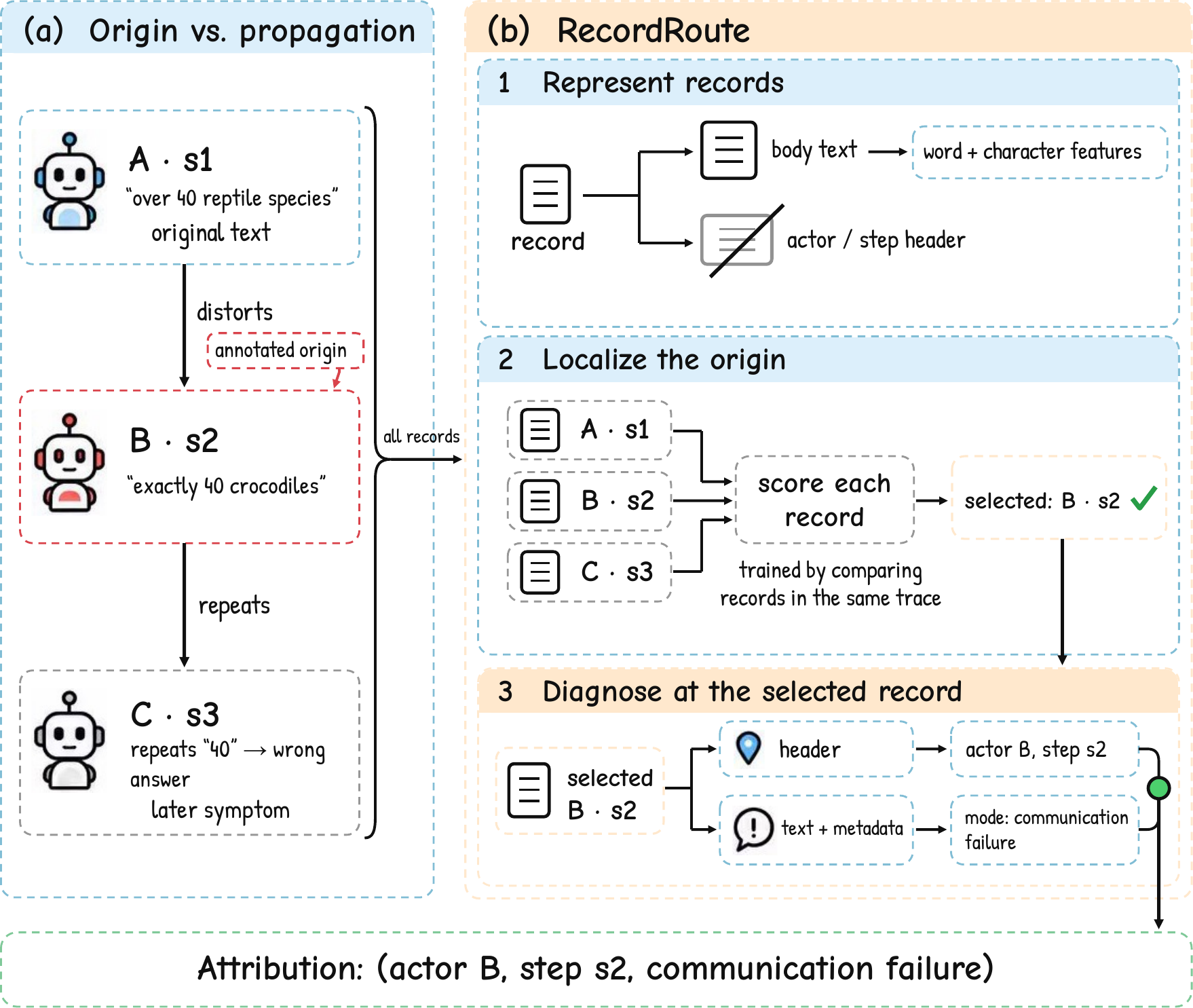}%
}
\newcommand{\generalizationgraphic}{%
  \includegraphics[width=\columnwidth]{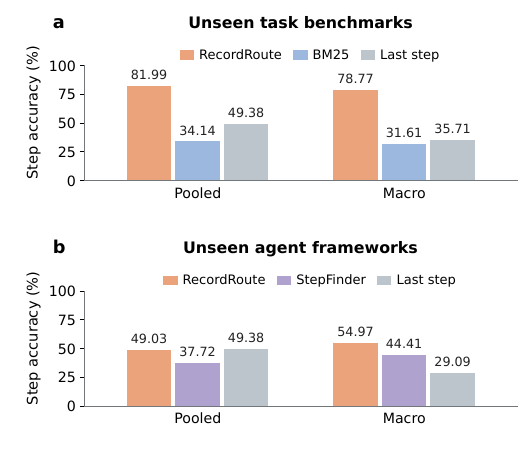}%
}

\title{RecordRoute: Auditing Agent Failures with Localization-Guided Diagnosis}

\author{
  Yifan Duan, Lihuan Li, Chao Xue, Siyuan Zheng, Du Yin, Flora D. Salim\thanks{Corresponding author.} \\
  UNSW Sydney, Australia \\
  \texttt{yifan.duan3@unsw.edu.au}, \texttt{flora.salim@unsw.edu.au}
}

\begin{document}
\begin{NoHyper}
\maketitle
\end{NoHyper}
\thispagestyle{arxivfirstpage}

\begin{abstract}
Auditing failed agent executions requires identifying the responsible
actor, the earliest decisive step, and the error mode. Representative
supervised methods either focus on step localization or predict multiple
attribution targets from shared trajectory representations. These methods
do not use the localization result to select the text for mode diagnosis.
To address this limitation, we propose RecordRoute, which couples comparative
localization with local diagnosis. Specifically, it represents each record
with word and character features and learns to rank records through
comparisons within each trajectory. The selected record supplies the actor and
step, while its text and metadata drive mode prediction. On
controlled-injection Who\&When Pro, RecordRoute outperforms StepFinder and
ASCon on all four official metrics. With actor and step predictions fixed,
local diagnosis reaches 77.47\% pooled joint accuracy versus 69.28\% for
full-trace diagnosis. On natural failures from TraceElephant and CatchBench
POST, it outperforms these baselines and three open LLMs in localization.
\end{abstract}

\section{Introduction}
\label{sec:introduction}

\begin{figure}[t]
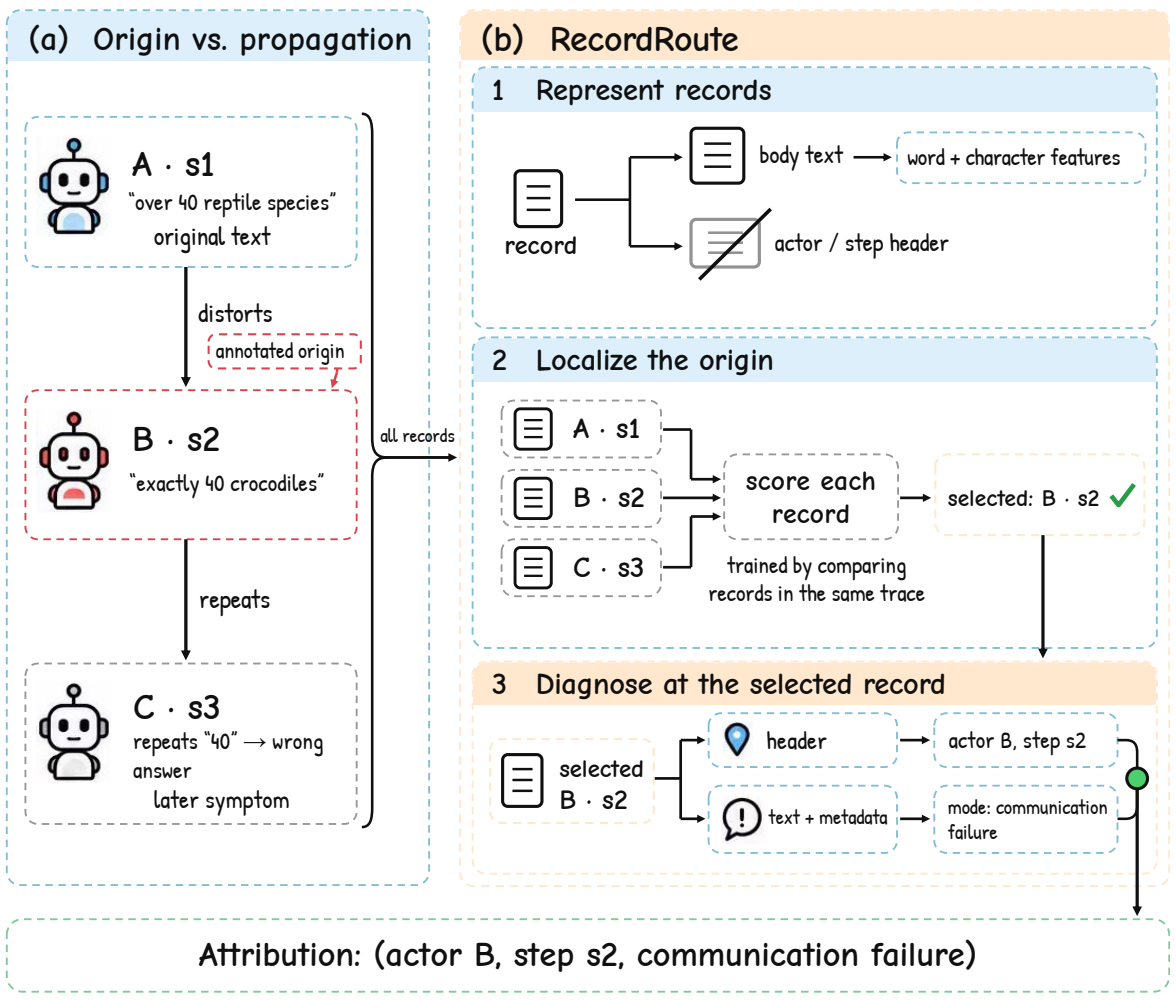

  \centering
  \traceexamplegraphic
  \caption{Overview of RecordRoute. (a) An error propagates across agents.
  (b) RecordRoute locates the actor and step, then diagnoses the error mode
  from the selected record.}
  \label{fig:trace-example}
\end{figure}

Language model agents increasingly execute long workflows using tools and
coordinating with other agents \citep{yao2023react,wu2024autogen}.
When such a workflow fails, the final answer reveals the outcome but rarely
identifies where the error began \citep{zhang2025which,barke2026agentrx}.
A useful audit requires identifying
the responsible actor, the earliest decisive step,
and the error mode \citep{zhang2025which,liu2026whowhenpro,chen2026seeing}.
Locating a later symptom instead of its origin directs debugging toward the
wrong component and obscures how the failure propagated
(Fig.~\ref{fig:trace-example}(a)).

Existing methods either ask a large language model to judge the trajectory
or score candidate steps with a smaller model. LLM-based methods read,
search, or verify the trajectory before generating a root-cause judgment
\citep{zhu2026raffles,xia2026agentlocate}. Some trace dependencies between
steps \citep{wang2026chief} or test hypotheses about the error
\citep{li2026errorprobe} to separate its origin from later effects.
MASPrism ranks steps
using token likelihoods and attention weights from a small language model,
without training \citep{liu2026masprism}. StepFinder and ASCon learn step
representations or joint agent and step representations from annotated
failures \citep{zhu2026stepfinder,jiang2026ascon}.

While promising, these approaches still struggle to separate a decisive
error from its later symptoms \citep{yang2026ceg}. First, evidence of an
earlier error can recur in later records \citep{pothuru2026failures}.
In Fig.~\ref{fig:trace-example}(a), agent C repeats agent B's incorrect
claim, so the error appears at both steps. Tool names, exception fragments,
and action wording can help distinguish such records. StepFinder and ASCon
represent steps with dense embeddings. Dense representations can struggle
with fine-grained lexical matching \citep{zhan2025lexsembridge}. Second,
these supervised models do not use the predicted step to select the text
for mode diagnosis. StepFinder focuses on step localization, while ASCon
predicts multiple attribution targets from shared trajectory
representations. Yet actor and step identify a record, and mode describes
the error in that record. A classifier that aggregates the full trace can
instead mix the decisive error with its later consequences. These
observations motivate comparing records within a trajectory while
preserving their literal details, and diagnosing the error from the
selected record.

To this end, we propose RecordRoute, a supervised method for locating and
diagnosing agent failures. After removing identifying headers, it represents each
record with word and character features. A pairwise ranker learns from
feature differences between records in the same training trajectory.
At inference, it scores records individually and selects the highest-scoring
one. The selected record supplies the actor and step predictions. Its text,
structural metadata, and actor identity provide the input for mode diagnosis
(Fig.~\ref{fig:trace-example}(b)).

On controlled-injection Who\&When Pro, RecordRoute outperforms StepFinder
and ASCon on all four official metrics.
It reaches 90.86\% pooled exact-step accuracy and 77.47\% pooled joint accuracy.
With actor and step predictions fixed, diagnosis from the selected record
improves joint accuracy by 8.18 percentage points over full-trace diagnosis.
The selected record contains a median of 19.77\% of preprocessed trace tokens.
On natural failures from TraceElephant and CatchBench POST, it ranks first
in localization (Table~\ref{tab:main-results}).

This work makes three contributions.
\begin{itemize}
    \setlength{\itemsep}{0pt}\setlength{\parsep}{0pt}
    \item We develop a locator that combines word and character features
    with pairwise ranking within each trajectory, preserving record details
    while learning to identify the responsible actor and step.
    \item We condition mode diagnosis on the selected record's text,
    structural metadata, and actor identity. The localization result determines
    the evidence used to diagnose the error.
    \item Ablations show that record representation, pairwise training, and
    local diagnosis each improve attribution on Who\&When Pro. RecordRoute
    outperforms StepFinder and ASCon across all three benchmarks.
\end{itemize}

\section{Related work}
\label{sec:related-work}

\paragraph{Failure-attribution benchmarks.}
Failure taxonomies and attribution benchmarks cover coordination errors,
responsible actors, decisive steps, and error types
\citep{cemri2025mast,zhang2025which,liu2026whowhenpro,kong2026aegis}.
Natural-failure and auditing benchmarks differ in what an attributor can
observe and how predictions are scored
\citep{chen2026seeing,barke2026agentrx,zhao2026catchbench}, while MP-Bench
and LongRCA study ambiguous attributions and root causes in long
trajectories \citep{in2026mpbench,zhang2026longrca}. We train on the labels
of each benchmark. The injected failures in Who\&When Pro support component
analysis, and the natural failures in TraceElephant and CatchBench test
localization.

\paragraph{Trace reasoning and intervention.}
Reasoning-based methods use multiple evaluators or scope narrowing
\citep{zhu2026raffles,xia2026agentlocate,sun2026scope}, dependency-guided
search \citep{wang2026chief,rafi2026falat,bakish2026adaptive,deng2026defa},
constraint or hypothesis verification
\citep{barke2026agentrx,li2026errorprobe,qiao2026verifymas,bi2026diagnosing},
and long-trace investigation \citep{zhu2026safari,qi2026trajdebug}.
AgentRx also evaluates a variant that first predicts the critical step and
then categorizes that step \citep{barke2026agentrx}. Other work filters
anomalous steps before attribution, uses attribution to guide reflection,
or maps failures to editable system assets
\citep{zhang2026duotrace,wang2026doctor,yuan2026daaf}. RecordRoute instead
learns a record ranker from annotated failures and diagnoses the error from
the selected record without generating text.

\paragraph{Learned and efficient attribution.}
StepFinder learns temporal step features. ASCon models reciprocal
context between agents and steps with target-specific heads
\citep{zhu2026stepfinder,jiang2026ascon}. Other approaches exploit graphs,
token likelihoods and attention weights from a small language model, or
successful trajectories
\citep{li2026afanet,zhao2026grade,liu2026masprism,yeh2026oat}.
AgenTracer post-trains a judge, HPFA contrasts failed and successful paths,
and conformal attribution returns calibrated error regions
\citep{zhang2025agentracer,zhu2026hpfa,feng2026conformal}. Unlike StepFinder
and ASCon, RecordRoute represents records with word and character features
and diagnoses the error from the selected record.

\section{Problem formulation}
\label{sec:task}

Consider an LLM-based agent system with actor set
$\mathcal A=\{a^{(1)},\ldots,a^{(N_a)}\}$. Its completed execution is recorded
as an ordered trajectory $T=(s_1,\ldots,s_{n_T})$, where each record has the
form
\begin{equation}
 s_i=(c_i,a_i,x_i,z_i), \qquad a_i\in\mathcal A.
 \label{eq:record-definition}
\end{equation}
Here $c_i$ is the benchmark's step or round coordinate, $a_i$ identifies the
executing actor, $x_i$ contains messages, reasoning text, or tool interactions,
and $z_i$ contains structural metadata such as record type and position.
The sequence index $i$ orders the records, while $c_i$ preserves the
benchmark coordinate. Several records can share $c_i$ when a round contains
actions by multiple actors. We denote the observed coordinate set by
$\mathcal C_T=\{c_i\}_{i=1}^{n_T}$ and the available task query by $q$.

Given a failed trajectory, attribution identifies the responsible actor,
the earliest decisive step, and its error mode. Let
$\mathcal M=\{m_1,\ldots,m_{N_m}\}$ denote the benchmark's error-mode taxonomy.
The annotation is represented by $\mathcal G_T=(A^*,C^*,m^*)$, where
$A^*\subseteq\mathcal A$ contains accepted responsible actors,
$C^*\subseteq\mathcal C_T$ contains accepted coordinates for the decisive
error, and $m^*\in\mathcal M$ is its mode label.

The objective is to learn an attribution function
\begin{equation}
 \begin{aligned}
 f_\theta(T,q)&=(\hat a,\hat c,\hat m),\\
 (\hat a,\hat c,\hat m)&\in\mathcal A\times\mathcal C_T\times\mathcal M.
 \end{aligned}
 \label{eq:attribution-function}
\end{equation}
The predicted actor and coordinate specify where the error is attributed,
and the predicted mode describes its type. A correct
attribution satisfies $\hat a\in A^*$, $\hat c\in C^*$, and $\hat m=m^*$,
where actor matches follow the official rules of each benchmark.
On TraceElephant and CatchBench, we evaluate step localization.

\section{Methodology}
\label{sec:method}

RecordRoute extracts features for each record, ranks the candidates, and
derives the attribution from the selected record. Unlike ASCon, which predicts modes
from an agent representation and pooled representations of its steps
\citep{jiang2026ascon}, our mode classifier reads the text of the record
selected by the locator. We represent records with word and character
features and train a locator $L_w$ through comparisons within each
trajectory. The selected record supplies the actor and step, and a
classifier $H_U$ predicts its mode, which decomposes
Equation~\ref{eq:attribution-function} as
\begin{equation}
 \begin{aligned}
 \hat i &= L_w(T,q),\\
 f_\theta(T,q) &=
 (a_{\hat i},c_{\hat i},H_U(s_{\hat i},q)).
 \end{aligned}
 \label{eq:method-decomposition}
\end{equation}
\begin{figure*}[t]
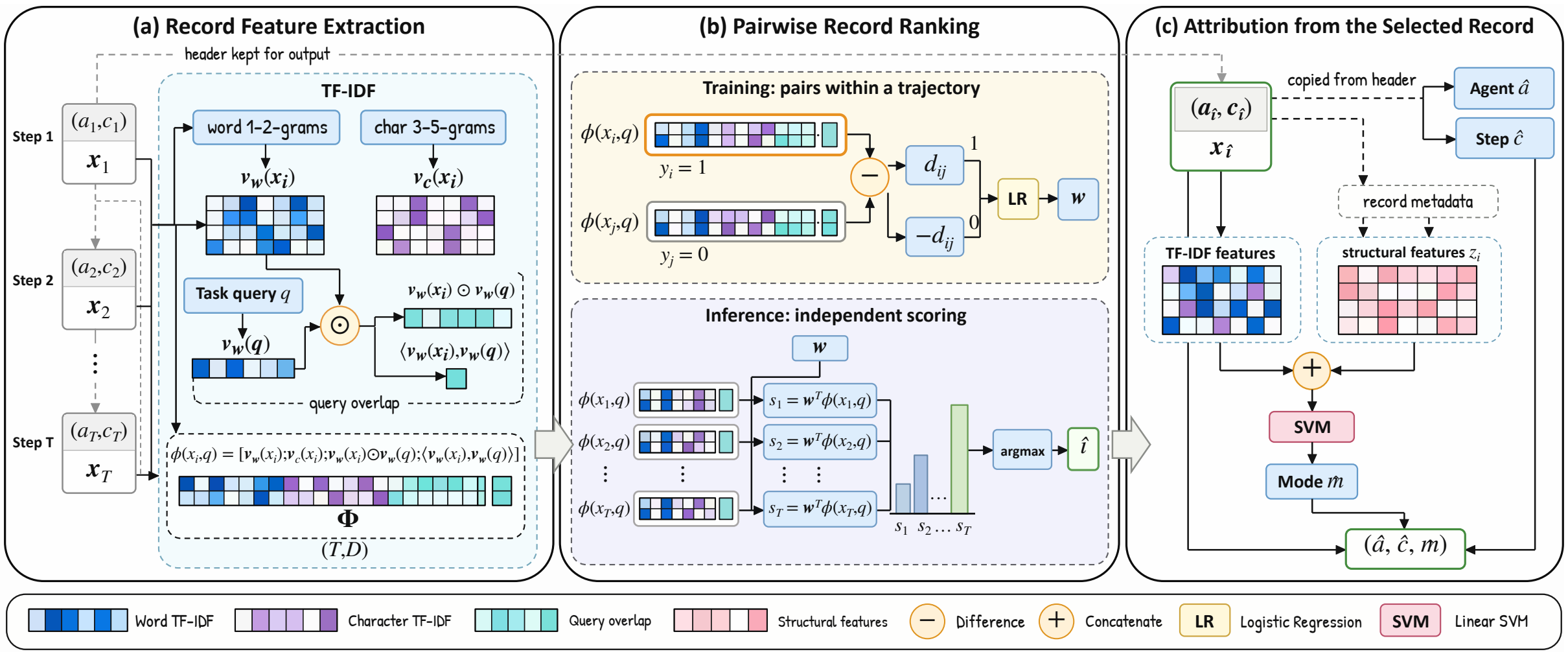

  \centering
  \methodoverviewgraphic
  \caption{Overview of RecordRoute. (a) Record features combine word and
  character TF-IDF with query overlap. Headers are retained for output.
  (b) Logistic regression learns from record pairs within a trajectory.
  At inference, the learned weights score records individually.
  (c) The selected record supplies actor and step from its retained header,
  and a linear SVM predicts the mode from its text and metadata.
  In the diagram, $T$ is the record count and $s_i$ is the score $\rho_i$.}
  \label{fig:method-overview}
\end{figure*}

\subsection{Record feature extraction}

Tool names, exception fragments, and action wording can help distinguish a
decisive error from records that repeat its symptoms. We retain these cues
with word and character TF-IDF features. Word features represent individual
terms and short action phrases. Character features retain substrings within
tool identifiers and error messages, so records can share features even when
their complete tokens differ. Together, the two views expose action wording
and local string details to the locator.

Before encoding a record, we remove headers that identify its actor and
coordinate. This prevents the rendered headers from contributing lexical
features to localization. The original actor and coordinate remain attached
to the candidate for output after localization. Each record remains a
separate candidate, preserving the correspondence between its text, actor,
and coordinate throughout encoding.

For record body $x_i$ and task query $q$, we concatenate four feature blocks
(Fig.~\ref{fig:method-overview}(a))
\begin{equation}
 \begin{aligned}
 \phi(x_i,q) = [
 &v_w(x_i),v_c(x_i),\\
 &v_w(x_i)\odot v_w(q),\\
 &\langle v_w(x_i),v_w(q)\rangle].
 \end{aligned}
 \label{eq:features}
\end{equation}
$v_w$ uses word unigrams and bigrams, and $v_c$ uses character n-grams of
lengths three to five within word boundaries. Elementwise multiplication
$\odot$ retains which terms are shared with the query, allowing the ranker
to assign different weights to individual matches. The inner product gives
an overall measure of lexical overlap. These two blocks connect record
wording to the task description while retaining the record's own word and
character features. When queries are unavailable or omitted, we set $q$ to
the empty string and both overlap blocks become zero. The resulting representation
$\phi_i=\phi(x_i,q)$ is the input to the locator.

\subsection{Pairwise record ranking}

Localization requires distinguishing the decisive error from other records
in the same trajectory. We train the locator to rank records accepted by the
annotation above the remaining candidates,
then score each record individually at inference
(Fig.~\ref{fig:method-overview}(b)). We use a coordinate target or a joint
actor and coordinate target, indexed by
$r\in\{\mathrm{coord},\mathrm{joint}\}$, to define candidate labels
\begin{equation}
 y_i^{(r)} =
 \begin{cases}
 \mathbf{1}[c_i\in C^*], & r=\mathrm{coord},\\
 \mathbf{1}[a_i\in A^*\wedge c_i\in C^*], & r=\mathrm{joint}.
 \end{cases}
 \label{eq:schema-target}
\end{equation}
Actor membership follows the official matching rules in
Section~\ref{sec:task}.
Coordinate targets apply to fine-grained records, while joint targets distinguish
actors within a shared round. The public trace schema determines the target
before model fitting. For each target, we fit a separate locator and mode
classifier on the same training split, and write $w$ and $U$ for the
parameters of the applicable variant.

For example, suppose actors A and B both act in round $c$, and A's record is
annotated as the decisive error. A coordinate target accepts both records at
$c$. A joint target accepts A's record and makes B's record a comparison
candidate. This allows the training labels to distinguish two actors even
when their records share the same step coordinate.

StepFinder uses a cross-entropy loss over candidate steps for localization
\citep{zhu2026stepfinder}. We instead train on feature differences between
pairs of records with different labels. For a fixed target, write
$y_i=y_i^{(r)}$ and let
$\mathcal P_T=\{(i,j)\mid 1\leq i,j\leq n_T,\ y_i\ne y_j\}$
contain all ordered pairs with different labels in each training trajectory.
Each informative pair appears in both orientations, with feature
differences and signed preference labels
\begin{equation}
 \begin{aligned}
 d_{ij}&=\phi_i-\phi_j,\\
 \ell_{ij}&=2\mathbf{1}[y_i>y_j]-1.
 \end{aligned}
 \label{eq:pairwise}
\end{equation}
These pairs compare the annotated error with other observations of the same
failed execution, including later symptoms. When several records are
accepted, each is compared with the remaining candidates, without imposing
an ordering among accepted records. For a linear score
$\rho_i=w^\top\phi_i$, the score difference satisfies
\begin{equation}
 w^\top d_{ij}=\rho_i-\rho_j.
 \label{eq:score-difference}
\end{equation}
Thus, training on feature differences directly learns the score ordering
used for selection. Features with equal values in both records cancel, so
the comparison depends on the differences between the two candidates.

Over the training trajectories $\mathcal D_{\mathrm{tr}}^+$ with nonempty
$\mathcal P_T$, we minimize the regularized pairwise logistic
loss \citep{burges2005ranknet}
\begin{equation}
 \begin{aligned}
 \mathcal L_{\mathrm{rank}}(w)
 &=\frac{\lambda}{2}\|w\|_2^2\\
 &\quad+\sum_{T\in\mathcal D_{\mathrm{tr}}^+}
 \frac{1}{|\mathcal P_T|}\sum_{(i,j)\in\mathcal P_T}\\
 &\qquad\log\big(1+\exp(-\ell_{ij}w^\top d_{ij})\big).
 \end{aligned}
 \label{eq:ranking-loss}
\end{equation}
Here $\lambda$ controls regularization. The loss encourages accepted
records to receive higher scores than other candidates in the same
trajectory. The factor $1/|\mathcal P_T|$ normalizes pair weights within each
informative trajectory to sum to one, which keeps trajectories with more
candidates from dominating the training loss. We fit this objective with
linear logistic regression. A shared intercept cancels
when subtracting two record scores, so the ranker is fitted without an intercept.

At inference, the locator scores each record once and selects the record
with the highest score
\begin{equation}
 \begin{aligned}
 \rho_i &= w^\top\phi_i,\\
 \hat i &= \arg\max_{1\leq i\leq n_T}\rho_i.
 \end{aligned}
 \label{eq:locator}
\end{equation}
We resolve score ties in favor of the earlier record. The selected index
$\hat i$ identifies the record passed to the attribution stage. Pair
construction is used only during training. At inference, the number of
score evaluations grows linearly with the number of records, followed by
one mode prediction.

\subsection{Attribution from the selected record}

Attribution begins by reading the actor and coordinate from the selected
record's retained header, giving
$(\hat a,\hat c)=(a_{\hat i},c_{\hat i})$
(Fig.~\ref{fig:method-overview}(c)). These predictions refer to the same
execution record even when several actors share a round coordinate.

The remaining question is which error mode describes the selected action.
A failed trajectory may contain the initiating error,
reactions to that error, and actions that are themselves valid. Combining
their text gives the classifier evidence about several events, although
the requested mode describes the error in the selected record. We
therefore use the selected record as the textual input to diagnosis. This
connects the predicted mode to the same execution record that supplies the
actor and coordinate.

The record's actor identity and structural metadata provide context for
interpreting its action. We encode the record text with the available task
query and concatenate the resulting features with its actor identity and structural
metadata
\begin{equation}
 h_i=[\widetilde\phi(x_i,q),\eta(a_i,z_i)].
 \label{eq:diagnostic-features}
\end{equation}
$\widetilde\phi$ uses the TF-IDF construction in
Equation~\ref{eq:features} and is fitted separately on records at accepted
coordinates, indexed by
$\mathcal I_T^{\mathrm{mode}}=\{i\mid c_i\in C^*\}$.
The metadata map $\eta$ combines standardized numeric features, including
record length, trajectory length, and relative position, with indicators
for record type and actor identity and their interactions with framework
and task domain. Local diagnosis thus restricts the execution text
to the selected record while retaining task and structural context.

A class-balanced linear SVM assigns a margin to each error mode. With
classifier parameters $U=\{(u_m,b_m)\}_{m\in\mathcal M}$, the prediction is
\begin{equation}
 \begin{aligned}
 \hat m&=H_U(s_{\hat i},q)\\
 &=\arg\max_{m\in\mathcal M}(u_m^\top h_{\hat i}+b_m).
 \end{aligned}
 \label{eq:mode-prediction}
\end{equation}
During training, we fit $U$ to examples $(h_i,m^*)$ for
$i\in\mathcal I_T^{\mathrm{mode}}$, with per-record weights
$1/|\mathcal I_T^{\mathrm{mode}}|$. Normalizing these weights prevents
trajectories with more records at accepted coordinates from receiving extra
sample weight. Restricting the training examples to these records associates
each mode label with its annotated coordinates. The locator and diagnostic classifier are fitted
separately using the same training split. At test time, the classifier
diagnoses the record selected by the locator, so localization errors remain
visible in joint attribution.

All feature maps and models are fitted using training data alone.
For step localization, we fit and apply the locator alone.
Preprocessing and model settings appear in
Appendix~\ref{sec:appendix-reproducibility}.

\section{Experiments}
\label{sec:experiments}

\subsection{Experimental Setup}

\paragraph{Datasets and Benchmarks.}

Our primary evaluation uses all 6,257 failures with text traces released by
Who\&When Pro \citep{liu2026whowhenpro}.  They contain 38,140 candidate
records from eight agent frameworks and 16 task benchmarks.
TraceElephant \citep{chen2026seeing} contains 218 natural failures and 5,898
candidate steps from three agent systems.
CatchBench \citep{zhao2026catchbench} defines PRE, LIVE, and POST audit
settings. Our method performs retrospective localization, which corresponds
to POST. We use its Who\&When subset, containing 126 natural failures and
1,099 candidate steps.

\paragraph{Baselines.}

We compare with StepFinder \citep{zhu2026stepfinder} and ASCon
\citep{jiang2026ascon} on all three benchmarks.
We use their official architectures and training budgets. Both use our mode
classifier on Who\&When Pro. ASCon retains its independent actor and step heads.
We also evaluate GRADE \texttt{exec-rank}, \texttt{auditable} blast share,
and PyGOD DOMINANT on all three benchmarks, adapting the CatchBench POST
implementations \citep{zhao2026catchbench}. These step rankers use the same
mode classifier on Who\&When Pro.
On CatchBench, we report the published POST results for
\texttt{auditable} and DOMINANT.
Simple baselines include random selection, BM25, selection of the first,
middle, or last step, and a learned relative-position baseline on all three
benchmarks.
We evaluate Qwen2.5-14B-Instruct \citep{qwen2025qwen25},
Mixtral-8x7B-Instruct-v0.1 \citep{jiang2024mixtral}, and Llama-3.1-8B-Instruct
\citep{grattafiori2024llama3} without benchmark-specific training on the
complete evaluation sets of all three benchmarks. Their prompts and input
settings appear in Appendix~\ref{sec:appendix-open-llms}.

\paragraph{Evaluation Metrics.}

On Who\&When Pro, we report pooled exact accuracy for each target and for
the complete tuple using the annotation-matching criteria in
Section~\ref{sec:task}. Joint accuracy requires all three targets to be correct.
We also report the official aggregates. Who is actor accuracy macro-averaged
over multi-agent frameworks, When is exact-step accuracy macro-averaged over
frameworks, What is global macro-F1 over the 14 observed error-mode classes,
and All is joint accuracy macro-averaged over frameworks.
TraceElephant uses exact-step accuracy. CatchBench uses Top-1 and Top-3
accuracy and mean reciprocal rank (MRR).

\paragraph{Implementation Details.}

We use five-fold cross-validation, grouping Who\&When Pro and TraceElephant
by task query and following CatchBench's official grouped splits.
Supervised methods use the same folds within each benchmark.
On TraceElephant, we select $C$ by cross-validation within each training fold.
For supervised methods, Who\&When Pro and TraceElephant record bodies are limited to 800 characters,
while CatchBench retains the original step length.
On Who\&When Pro, StepFinder and ASCon use three training seeds.
Our CatchBench results average five split seeds.
Preprocessing, training settings, and aggregation details appear in
Appendix~\ref{sec:appendix-reproducibility}.

\subsection{Main Results}
\label{sec:results}

\begin{table*}[t]
\centering
\small
\setlength{\tabcolsep}{4pt}
\setlength{\dashlinedash}{3pt}
\setlength{\dashlinegap}{2pt}
\setlength{\arrayrulewidth}{0.4pt}
\begin{tabular*}{\textwidth}{@{\extracolsep{\fill}}lrrrrrrrr@{}}
\toprule
& \multicolumn{4}{c}{Who\&When Pro}
& \multicolumn{1}{c}{TraceElephant}
& \multicolumn{3}{c}{CatchBench POST} \\
\cmidrule(lr){2-5}\cmidrule(lr){6-6}\cmidrule(lr){7-9}
Method & Who & When & What & All & Step & Top-1 & Top-3 & MRR \\
\midrule
Random & 41.04 & 31.11 & 27.63 & 17.77 & 8.72 & 11.90 & 34.60 & 0.324 \\
BM25 & 49.59 & 38.05 & 25.53 & 20.26 & 7.80 & 12.70 & 65.87 & 0.420 \\
First-step & 56.97 & 34.83 & 13.43 & 12.18 & 5.50 & 15.87 & 51.59 & 0.407 \\
Middle-step & 36.06 & 24.96 & 18.31 & 11.44 & 8.72 & 7.94 & 42.06 & 0.330 \\
Last-step & 31.68 & 29.09 & 31.63 & 24.83 & 2.29 & 0.79 & 18.25 & 0.225 \\
Position & 31.68 & 29.09 & 31.63 & 24.83 & 5.50 & 15.87 & 51.59 & 0.407 \\
\noalign{\vskip\aboverulesep}\hdashline\noalign{\vskip\belowrulesep}
StepFinder & 62.05 & 73.28 & 56.62 & 48.53 & 11.01 & 17.46 & 48.41 & 0.394 \\
ASCon & 70.28 & 77.06 & 60.06 & 52.30 & 18.81 & 34.92 & 72.54 & 0.567 \\
\noalign{\vskip\aboverulesep}\hdashline\noalign{\vskip\belowrulesep}
\texttt{auditable} & 56.97 & 34.83 & 13.43 & 12.18 & 5.50 & 15.9 & 51.6 & 0.407 \\
PyGOD DOMINANT & 43.71 & 42.52 & 30.12 & 22.42 & 5.50 & 4.8 & 30.2 & 0.258 \\
GRADE exec-rank & 52.45 & 45.96 & 34.09 & 32.88 & 8.72 & 21.11 & 61.43 & 0.454 \\
\noalign{\vskip\aboverulesep}\hdashline\noalign{\vskip\belowrulesep}
Qwen2.5-14B-Instruct & 58.52 & 62.34 & 14.05 & 15.91 & 23.85 & 18.25 & 53.17 & 0.421 \\
Mixtral-8x7B-Instruct & 53.10 & 45.90 & 9.62 & 9.26 & 11.47 & 19.84 & 51.59 & 0.424 \\
Llama-3.1-8B-Instruct & 48.13 & 43.72 & 6.27 & 8.12 & 12.84 & 7.94 & 43.65 & 0.331 \\
\noalign{\vskip\aboverulesep}\hdashline\noalign{\vskip\belowrulesep}
\textbf{RecordRoute (ours)} & \textbf{85.14} & \textbf{88.50} & \textbf{70.05} & \textbf{63.67} & \textbf{27.06} & \textbf{37.46} & \textbf{77.78} & \textbf{0.590} \\
\bottomrule
\end{tabular*}
\caption{Main results on three benchmarks. Scores are percentages except MRR.
All three open LLMs cover the full evaluation sets without benchmark training labels.
Prompt and decoding settings appear in Appendix~\ref{sec:appendix-open-llms}.}
\label{tab:main-results}
\end{table*}

Table~\ref{tab:main-results} compares methods across three benchmarks.
RecordRoute learns from labeled failures, while the three open LLMs
receive no benchmark training labels.

\paragraph{Who\&When Pro.}
RecordRoute achieves the highest scores on all four attribution metrics.
It reaches 88.50 When and 63.67 All, compared with 73.28 and 48.53 for
StepFinder, and 77.06 and 52.30 for ASCon.
The All gains of 15.14 and 11.37 percentage points measure improvements
in joint actor, step, and mode prediction. RecordRoute also improves
Who and What over ASCon by 14.86 and 9.99 points.
Both neural baselines share our mode classifier, keeping the diagnosis
model fixed while the selected records differ.
The supervised text-based methods also exceed the structural rankers in
When and All. GRADE \texttt{exec-rank}, the strongest structural baseline,
reaches 45.96 When and 32.88 All.
Qwen2.5-14B leads the three open LLMs with 62.34 When and 15.91 All.
Qwen scores 14.05 on What, compared with 70.05 for RecordRoute, indicating
that the gap includes diagnosis of the error mode.
RecordRoute's gains over this zero-shot baseline cover both localization
and complete attribution.

\paragraph{TraceElephant.}
The localization advantage also appears on naturally occurring failures.
We train and evaluate each supervised method on TraceElephant using identical
folds grouped by task. RecordRoute reaches 27.06\% exact-step accuracy,
compared with 18.81\% for ASCon and 11.01\% for StepFinder.
The gains are 8.26 and 16.06 percentage points, respectively.
Qwen2.5-14B, the strongest evaluated open LLM on this benchmark, reaches
23.85\% with the official prompt.
BM25 reaches 7.80\%, while GRADE \texttt{exec-rank} reaches 8.72\%.
Both neural baselines improve on retrieval and execution features, and
RecordRoute improves on both neural baselines.
This ordering among the supervised methods is consistent with
Who\&When Pro despite the different agent systems and source of failures.
Task grouping across systems excludes each test query from the
corresponding training folds.

\paragraph{CatchBench.}
RecordRoute achieves the highest mean Top-1, Top-3, and MRR scores among
all methods in Table~\ref{tab:main-results} on CatchBench POST.
It reaches 37.46\% Top-1, 77.78\% Top-3, and 0.590 MRR.
ASCon is the closest supervised competitor, with 34.92\% Top-1,
72.54\% Top-3, and 0.567 MRR.
The open LLMs also score lower when asked to rank all candidates.
Mixtral leads them in Top-1 and MRR at 19.84\% and 0.424, while Qwen
reaches 53.17\% Top-3.
The comparison with BM25 illustrates why both first-choice accuracy and
candidate coverage matter. BM25 reaches 65.87\% Top-3 but only
12.70\% Top-1, often including the fault without placing it first.
RecordRoute improves both measures and the overall ranking measured by
MRR. Its top three candidates contain the annotated fault in over
three quarters of cases, providing a short list for subsequent inspection.
Pooled scores and results across training seeds appear in
Appendix~\ref{sec:appendix-baselines}.

\subsection{Generalization}
\label{sec:generalization}

\begin{figure}[t]
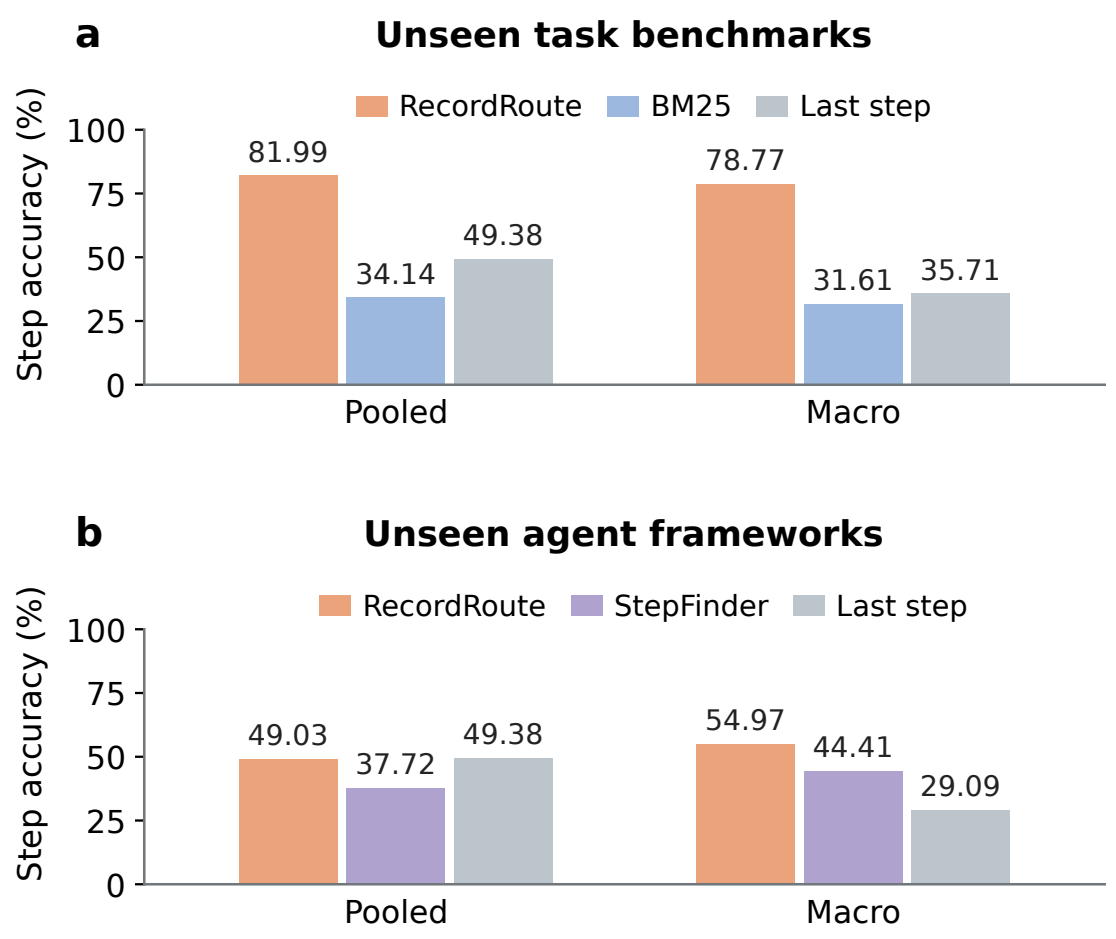

\centering
\generalizationgraphic
\caption{Exact-step accuracy on unseen (a) task benchmarks and (b) agent
frameworks. Pooled weights trajectories equally, while macro weights
benchmarks or frameworks equally.}
\label{fig:generalization}
\end{figure}

We evaluate whether RecordRoute can localize errors in task benchmarks and
agent frameworks excluded from training. For each task benchmark,
we train the locator on the other 15 benchmarks after removing matching task
queries. The locator receives only record text and achieves 81.99\% pooled
and 78.77\% macro exact-step accuracy (Figure~\ref{fig:generalization}(a)),
outperforming BM25 and the last-step baseline on all 16 benchmarks.
These results show that the locator can identify errors on new task benchmarks
using the execution records alone, without access to the task query.

We further test whether this localization capability extends to unseen
agent frameworks by excluding each framework and its matching task queries
from training. RecordRoute outperforms StepFinder on seven of the eight
frameworks and reaches 54.97\% macro exact-step accuracy, compared with
44.41\% for StepFinder (Figure~\ref{fig:generalization}(b)). The gains across
these frameworks show that the learned ranking remains useful when the
agent framework changes. Pooled accuracy is 49.03\%, compared with 49.38\%
for the last-step baseline.

\subsection{Ablation Study}

\subsubsection{Record representation and ranking}
\label{sec:results-components}

\begin{table}[t]
\centering
\small
\setlength{\tabcolsep}{3pt}
\begin{tabular*}{\columnwidth}{@{\extracolsep{\fill}}lrrrr@{}}
\toprule
\multicolumn{5}{@{}l}{\textit{(a) Who\&When Pro}} \\
& \multicolumn{2}{c}{Pointwise} & \multicolumn{2}{c}{Pairwise} \\
\cmidrule(lr){2-3}\cmidrule(l){4-5}
Representation & Step & Joint & Step & Joint \\
\midrule
Qwen3 (128d) & 61.36 & 52.76 & 65.10 & 56.10 \\
Qwen3 (1024d) & 69.63 & 60.32 & 72.69 & 62.92 \\
\shortstack[l]{Word and character\\TF-IDF}
& \textbf{86.73} & \textbf{74.06} & \textbf{90.41} & \textbf{77.13} \\
\end{tabular*}\par
\begin{tabular*}{\columnwidth}{@{\extracolsep{\fill}}lrrr@{}}
\midrule
\multicolumn{4}{@{}l}{\textit{(b) CatchBench POST}} \\
Objective & Top-1 & Top-3 & MRR \\
\midrule
Pointwise & 34.44 & 73.97 & 0.570 \\
Pairwise & \textbf{37.46} & \textbf{77.78} & \textbf{0.590} \\
\bottomrule
\end{tabular*}
\caption{Record representations and ranking objectives. Accuracies are
percentages, pooled across trajectories for Who\&When Pro.}
\label{tab:representation-objective}
\end{table}

We compare sparse record features with Qwen3 embeddings at 128 and 1024
dimensions under pointwise and pairwise learning. All six configurations
use the same folds, step targets, and mode classifier, with blank task queries
and $C=1$.

Table~\ref{tab:representation-objective}(a) identifies two sources of
localization accuracy. At the stronger 1024-dimensional dense control, the
word and character representation contributes 17.41 exact-step points.
Pairwise learning contributes another 3.36 points, with an interaction of
$+0.62$ points. At StepFinder's 128-dimensional setting, the corresponding
effects are 25.35 and 3.71 points. The sparse pairwise model exceeds the
1024-dimensional dense control by 17.72 points.
Its 90.41\% rises to the final 90.86\% with query overlap
and targets that distinguish actors within shared steps.

Word and character features lead under both objectives, while pairwise
learning improves each representation. The same ordering holds for joint
accuracy, reaching 77.13\% with sparse pairwise training versus 62.92\%
with 1024-dimensional embeddings under the same mode classifier.

On CatchBench, pairwise learning improves all three ranking metrics with
the same record features (Table~\ref{tab:representation-objective}(b)),
supporting comparative ranking on natural failures.
Top-1 accuracy rises from 34.44\% to 37.46\%.

With task queries removed and content words masked, the locator still
achieves 86.75\% exact-step accuracy on Who\&When Pro
(Table~\ref{tab:content-controls}(a)).

\subsubsection{Diagnostic evidence}
\label{sec:results-diagnosis}

\begin{table}[t]
\centering
\small
\setlength{\tabcolsep}{4.5pt}
\begin{tabular*}{\columnwidth}{@{\extracolsep{\fill}}lrr@{}}
\toprule
Variant & Mode & Joint \\
\midrule
\textbf{RecordRoute} & \textbf{78.87} & \textbf{77.47} \\
Without record text & 45.58 & 44.38 \\
\midrule
\multicolumn{3}{@{}l}{\textit{Full-trace diagnosis}} \\
Concatenation & 71.02 & 69.28 \\
Mean pooling & 46.84 & 43.92 \\
Max pooling & 55.17 & 52.63 \\
Learned aggregation & 68.24 & 64.95 \\
\cmidrule{1-3}
Separate actor prediction & 68.99 & 65.41 \\
\bottomrule
\end{tabular*}
\caption{Diagnosis ablations with the locator fixed. Mode and Joint are
pooled accuracies (\%).}
\label{tab:diagnostic-route}
\end{table}

With actor and step predictions fixed, we compare the selected record and
the concatenated full trace using the same classifier design, metadata,
folds, and regularization (Table~\ref{tab:diagnostic-route}). Diagnosis from
the selected record reaches 77.47\% joint accuracy, compared with 69.28\%
from the full trace, a gain of 8.18 points. This comparison supports using
the locator's selection to choose evidence for the mode classifier.

Mean and max pooling reach 43.92\% and 52.63\% joint accuracy.
These controls combine class margins from individual records, whereas
concatenation combines their text. Local diagnosis exceeds both forms of
aggregation, using only the record selected by the locator.

Learned aggregation of predictions from individual records reaches 64.95\%
joint accuracy, leaving a 12.51-point gap. The selected record therefore
remains more effective when the full-trace control processes records
individually before combining their predictions. With the same step
predictions, independent actor and full-trace mode classifiers reach
65.41\%. Removing text from the selected record while retaining its
structure and actor reduces joint accuracy to 44.38\%, showing that its
text contributes information beyond the metadata.

After both diagnostic models are retrained with content words masked, the
selected record remains better than learned full-trace aggregation, with
72.81\% versus 57.41\% joint accuracy
(Table~\ref{tab:content-controls}(b)).

On the 597 trajectories in which multiple actors share step coordinates,
defining positive records by both actor and step reduces their mean number
from 3.23 to 1.35. Twelve previously incorrect joint predictions become
correct and one becomes incorrect, with predictions for unique coordinates
unchanged.

With raw-text input, median inference takes 9.32\,ms on CPU for RecordRoute
and 66.61\,ms on GPU for StepFinder.
Efficiency measurements and diagnostic controls using LLMs appear in
Appendices~\ref{sec:appendix-reproducibility} and~\ref{sec:appendix-llm-diagnosis},
respectively.

\section{Conclusion}

We presented RecordRoute, a supervised method for identifying the responsible
actor, earliest decisive step, and error mode.
It combines word and character features with pairwise ranking within each
trajectory.
The selected record supplies the actor and step, while its text and metadata
guide mode prediction.
Experiments show gains over StepFinder and ASCon across three benchmarks,
with ablations supporting all three components.
These findings support using localization to select the evidence for diagnosis.

\bibliography{references}

\appendix
\section{Implementation Details}
\label{sec:appendix-reproducibility}

\paragraph{Data and preprocessing.}
We group Who\&When Pro and TraceElephant into five folds by the task query
normalized to lowercase with whitespace collapsed. Vocabularies, scalers,
and classifiers are fitted on the training folds. Record headers are
removed from the locator input. Record bodies are limited to 800
characters, retaining 58\% from the beginning and 42\% from the end.
TraceElephant inputs include outputs and linked tool results. CatchBench
uses its official grouped folds and five split seeds, retaining each
record's original length. Both natural-failure benchmarks use record text
without task queries. For the generalization experiments, training excludes
the evaluated task benchmark or framework and all matching task queries.

\paragraph{Features and classifiers.}
Word TF-IDF uses unigrams and bigrams with at most 30,000 features.
Character TF-IDF uses n-grams of lengths three to five within word boundaries
with at most 20,000 features. Both use sublinear term frequency and a minimum
document frequency of two. The pairwise logistic ranker has no intercept
and uses $C=1$ on Who\&When Pro and CatchBench. The class-balanced linear
SVM for mode prediction also uses $C=1$. On TraceElephant, five-fold
validation within each training fold selects $C$ from
$\{0.01,0.03,0.1,0.3,1,3,10\}$ by pooled exact-step accuracy, with ties
favoring the smaller value. The five selected values are $1,3,3,1,0.3$.

\paragraph{Baselines and seeds.}
StepFinder and ASCon use Qwen3-Embedding-0.6B
\citep{zhang2025qwen3embedding} with 128-dimensional record embeddings and
32-dimensional actor embeddings. StepFinder trains for up to 50 epochs
with batch size 16 and patience 10. ASCon uses its public Task-1 architecture
and 12-epoch schedule. Both use the same inner development split and
training seeds 42, 43, and 44 on Who\&When Pro, with the mode classifier
refitted in each fold. ASCon retains its independent actor head.
Natural-failure runs use training seed $42+\text{fold}$. CatchBench reserves
10\% of each training fold for checkpoint selection, and its results
average the five official split seeds. Framework evaluation uses seed 42
for StepFinder.

Random selection uses a fixed trace-ID hash. BM25 ranks records against
the task query. First and Last rank records chronologically and in reverse,
and Middle ranks by distance from the lower midpoint. Position uses
class-balanced logistic regression on relative position.
GRADE retains its eight public execution features and uses $C=1$.
The graph baselines assume full-context dependencies, under which
\texttt{auditable} ranks records chronologically. DOMINANT fits unlabeled
run graphs with two layers, hidden width 32, 40 epochs, and seed 0 on CPU.
Its feature scaling and graph fitting use the unlabeled dataset, while
its mode classifier uses only the corresponding training folds.

\paragraph{Diagnostic controls.}
The concatenation control reads the full trace. Mean and max pooling
combine class margins from individual records. Learned aggregation fits a
classifier on their means, maxima, and standard deviations using inner
out-of-fold predictions. The separate-actor control fits an actor ranker
on aggregated actor text and a mode classifier on the full trace.
These controls retain the same step predictions, feature family, folds,
and regularization as local diagnosis.

\paragraph{Efficiency.}
\label{sec:appendix-efficiency}
Table~\ref{tab:efficiency} reports measurements on 256 trajectories using
one RTX A5000 host. Timings include feature construction, localization,
and diagnosis, excluding model loading and disk I/O. After one warm-up,
we use ten cached-path repeats and three raw StepFinder repeats.
Training takes 159.75 seconds per fold for RecordRoute and 474.52 for
StepFinder.

\begin{table*}[t]
\centering
\small
\begin{tabular}{lllrrrr}
\toprule
Pipeline & Input & Device & ms/inc. & RAM (GiB) & State (MiB) & VRAM (GiB) \\
\midrule
RecordRoute & raw text & CPU & 9.32 & 1.66 & 12.75 & 0.00 \\
StepFinder & cached features & GPU & 2.21 & 1.81 & 11.48 & 0.04 \\
StepFinder & raw text & GPU & 66.61 & 1.92 & 1137.41 & 5.08 \\
\bottomrule
\end{tabular}
\caption{Median inference time and peak memory. Raw StepFinder input
includes embedding computation.}
\label{tab:efficiency}
\end{table*}

\section{LLM Experiments}
\label{sec:appendix-llm-experiments}

\subsection{Open LLM baselines}
\label{sec:appendix-open-llms}

Qwen2.5-14B-Instruct, Mixtral-8x7B-Instruct-v0.1, and Llama-3.1-8B-Instruct
receive no benchmark training examples. On Who\&When Pro, the prompt
supplies the official trace rendering, task query, and error taxonomy,
and requests the actor, step coordinate, and error mode. Explicit
coordinates and taxonomy names are normalized before official scoring.
Missing or unrecognized fields count as incorrect.

For Who\&When Pro and CatchBench, inference uses vLLM 0.26.0, BF16 weights,
greedy decoding, seed 0, a 30,000-token input limit, and up to 512 output
tokens. Longer inputs retain every record header and each body's beginning
and end under a common character cap. All cases remain in the evaluation.
CatchBench prompts request a ranking of all candidate step IDs. Incomplete
rankings receive a format retry, followed if needed by decoding constrained
to complete permutations of the candidate IDs. The same ranking determines
Top-1, Top-3, and MRR.

On TraceElephant, Qwen uses the official all-at-once prompt, while Mixtral
and Llama use the shared direct-attribution prompt. Inference uses BF16
weights, greedy decoding, and a 512-token output limit. All 218 cases are
scored, including the 57 Mixtral answers without a valid global step
coordinate, which count as incorrect.

\subsection{Diagnosis with selected records}
\label{sec:appendix-llm-diagnosis}

We evaluate Claude Sonnet 5.5 and DeepSeek V4.1 Flash on the same 128
Who\&When Pro cases, with 16 from each framework. The sample and prompts
are fixed before inference. Both models receive the task query and public
error taxonomy. Direct attribution requests all three targets from the
full trace. The RecordRoute conditions fix its predicted actor and step
and request the mode from either the selected record or the full trace.
BM25 uses the same local diagnosis prompt with its own selected record.
RecordRoute uses labeled training failures from the target frameworks,
while the LLMs receive no training examples.

We use \texttt{claude-sonnet-5-5} with high effort and
\texttt{deepseek-flash} with reasoning disabled. Each condition uses matching
instructions and output fields across providers. The output limit is 4,096
tokens, with one retry at 8,192 tokens for truncated answers. The calls
were made in October 2026, and every condition is scored on all 128 cases.

\begin{table*}[t]
\centering
\small
\caption{Pooled attribution accuracy (\%) on 128 Who\&When Pro cases.}
\label{tab:api-llm-diagnosis}
\setlength{\tabcolsep}{4pt}
\begin{tabular*}{\textwidth}{@{\extracolsep{\fill}}lrrrrrrrr@{}}
\toprule
 & \multicolumn{4}{c}{Claude Sonnet 5.5} & \multicolumn{4}{c}{DeepSeek V4.1 Flash} \\
\cmidrule(lr){2-5}\cmidrule(l){6-9}
Method & Actor & Step & Mode & Joint & Actor & Step & Mode & Joint \\
\midrule
LLM only & 74.22 & 67.19 & 27.34 & 19.53 & 75.00 & 66.41 & 31.25 & 20.31 \\
BM25 + local diagnosis & 60.94 & 41.41 & 21.09 & 10.94 & 60.94 & 41.41 & 23.44 & 8.59 \\
RecordRoute + full trace & 91.41 & 87.50 & 34.38 & 32.81 & 91.41 & 87.50 & 35.16 & 30.47 \\
RecordRoute + local diagnosis & 91.41 & 87.50 & \textbf{44.53} & \textbf{41.41} & 91.41 & 87.50 & \textbf{39.06} & \textbf{36.72} \\
\bottomrule
\end{tabular*}
\end{table*}

Table~\ref{tab:api-llm-diagnosis} shows that RecordRoute with local diagnosis
raises joint accuracy from 19.53\% to 41.41\% for Sonnet and from 20.31\%
to 36.72\% for DeepSeek. Both models also obtain higher mode accuracy from
the selected record than from the full trace at the same predicted
location. Local diagnosis reduces API input tokens by 54.86\% and 57.00\%,
respectively, relative to direct attribution. These token counts exclude
locator training and inference. Supplying predicted locations and selected
records improves both LLMs on this cohort. The original RecordRoute
classifier reaches 58.59\% joint accuracy on the same cases.

\section{Supplementary Results}
\label{sec:appendix-supplementary}
\label{sec:appendix-baselines}

\paragraph{Pooled results and training seeds.}
Tables~\ref{tab:whowhen-pooled} and~\ref{tab:baseline-seeds} report pooled
exact accuracy and individual training seeds on Who\&When Pro. Folds and
checkpoint rules are fixed across seeds.

\paragraph{Record content.}
Table~\ref{tab:content-controls} reports localization and diagnosis after
changes to record text. Code canonicalization standardizes Markdown fences
and \texttt{<code>} tags. Content masking replaces content words and numbers
while retaining punctuation, word lengths, and a fixed vocabulary of
function and heading words. Masked localization removes task queries.
Both diagnostic models are retrained on masked records with actor and
step predictions fixed.

\paragraph{Individual benchmarks and frameworks.}
\label{sec:appendix-generalization}
Tables~\ref{tab:lobo-benchmark} and~\ref{tab:lofo-step} give the individual
results underlying Figure~\ref{fig:generalization}.

\begin{table*}[t]
\begin{minipage}[t]{0.48\textwidth}
\vspace{0pt}
\centering
\small
\setlength{\tabcolsep}{1.5pt}
\begin{tabular}{lrr}
\toprule
Method & Step exact & Joint exact \\
\midrule
Stable random & 29.01 & 22.92 \\
BM25 & 34.14 & 27.33 \\
First-step & 14.38 & 5.74 \\
Middle-step & 16.56 & 10.76 \\
Last-step & 49.38 & 44.38 \\
Position & 49.38 & 44.38 \\
StepFinder & 75.53 & 63.61 \\
ASCon & 78.05 & 65.94 \\
\texttt{auditable} & 14.38 & 5.74 \\
PyGOD DOMINANT & 40.64 & 31.40 \\
GRADE exec-rank & 52.71 & 46.08 \\
Separate actor prediction & \textbf{90.86} & 65.41 \\
\textbf{RecordRoute (ours)} & \textbf{90.86} & \textbf{77.47} \\
\bottomrule
\end{tabular}
\caption{Pooled exact accuracy (\%) on Who\&When Pro. StepFinder and ASCon
average three seeds.}
\label{tab:whowhen-pooled}

\end{minipage}\hfill
\begin{minipage}[t]{0.48\textwidth}
\vspace{0pt}
\centering
\footnotesize
\setlength{\tabcolsep}{2pt}
\begin{tabular*}{\linewidth}{@{\extracolsep{\fill}}lrrrrrr@{}}
\toprule
Seed & Who & When & What & All & Step & Joint \\
\midrule
\multicolumn{7}{@{}l}{\textit{StepFinder}} \\
42 & 62.76 & 73.47 & 57.14 & 48.84 & 75.90 & 63.82 \\
43 & 61.02 & 72.32 & 55.65 & 47.19 & 74.84 & 62.83 \\
44 & 62.38 & 74.05 & 57.06 & 49.57 & 75.84 & 64.18 \\
\midrule
\multicolumn{7}{@{}l}{\textit{ASCon}} \\
42 & 70.17 & 77.36 & 60.35 & 51.82 & 78.20 & 65.78 \\
43 & 69.66 & 76.98 & 60.85 & 52.60 & 78.15 & 66.10 \\
44 & 71.00 & 76.85 & 58.98 & 52.46 & 77.80 & 65.94 \\
\bottomrule
\end{tabular*}
\caption{Who\&When Pro results (\%) across training seeds. Step and Joint
are pooled accuracies.}
\label{tab:baseline-seeds}

\end{minipage}
\end{table*}

\begin{table*}[t]
\centering
\begin{minipage}[t]{0.48\textwidth}
\footnotesize
\setlength{\tabcolsep}{2.5pt}
\begin{tabular*}{\linewidth}{@{\extracolsep{\fill}}lrrr@{}}
\toprule
\multicolumn{4}{@{}l}{\textit{(a) Localization}} \\
Input & All & SmolAgents & Others \\
\midrule
Original & 90.86 & 92.61 & 88.93 \\
Canonical code markup & 91.24 & 93.07 & 89.23 \\
Masked, no query & 86.75 & 89.74 & 83.46 \\
Last-step & 49.38 & -- & -- \\
\bottomrule
\end{tabular*}
\end{minipage}\hfill
\begin{minipage}[t]{0.48\textwidth}
\footnotesize
\setlength{\tabcolsep}{3pt}
\begin{tabular*}{\linewidth}{@{\extracolsep{\fill}}llrr@{}}
\toprule
\multicolumn{4}{@{}l}{\textit{(b) Diagnosis}} \\
Input & Evidence & Mode & Joint \\
\midrule
Original & Selected record & 78.87 & 77.47 \\
         & Learned aggregation & 68.24 & 64.95 \\
Masked & Selected record & 74.05 & 72.81 \\
       & Learned aggregation & 59.63 & 57.41 \\
\bottomrule
\end{tabular*}
\end{minipage}
\caption{Who\&When Pro accuracy (\%) under changes to record text.
(a) Exact-step localization. (b) Diagnosis with actor and step predictions fixed.}
\label{tab:content-controls}
\end{table*}

\begin{table*}[t]
\begin{minipage}[t]{0.48\textwidth}
\vspace{0pt}
\centering
\footnotesize
\setlength{\tabcolsep}{2.5pt}
\begin{tabular*}{\linewidth}{@{\extracolsep{\fill}}lrrrr@{}}
\toprule
Benchmark & $n$ & Ours & BM25 & Last \\
\midrule
ALFWorld & 311 & \textbf{25.40} & 5.47 & 0.00 \\
BigCodeBench & 195 & \textbf{55.90} & 25.13 & 0.00 \\
DABench & 327 & \textbf{97.55} & 40.37 & 52.29 \\
DACode & 306 & \textbf{78.10} & 11.11 & 11.11 \\
DataBench & 818 & \textbf{99.02} & 47.43 & 56.72 \\
DSQA & 830 & \textbf{65.78} & 11.45 & 57.47 \\
GAIA & 302 & \textbf{64.24} & 31.13 & 39.07 \\
GPQA & 143 & \textbf{88.81} & 65.03 & 15.38 \\
HumanEval & 18 & \textbf{94.44} & 11.11 & 0.00 \\
KramaBench & 82 & \textbf{97.56} & 15.85 & 14.63 \\
LiveCodeBench Pro & 129 & \textbf{41.86} & 39.53 & 0.00 \\
MATH & 448 & \textbf{93.75} & 33.48 & 74.78 \\
MMLU-Pro & 371 & \textbf{91.91} & 27.22 & 62.53 \\
SciBench & 555 & \textbf{89.01} & 38.56 & 61.26 \\
SimpleQA Verified & 509 & \textbf{78.00} & 58.55 & 66.01 \\
TableBench & 913 & \textbf{99.01} & 44.36 & 60.13 \\
\midrule
Pooled & 6257 & \textbf{81.99} & 34.14 & 49.38 \\
Benchmark macro & -- & \textbf{78.77} & 31.61 & 35.71 \\
\bottomrule
\end{tabular*}
\caption{Exact-step accuracy (\%) on new task benchmarks. The locator
receives no task query.}
\label{tab:lobo-benchmark}

\end{minipage}\hfill
\begin{minipage}[t]{0.48\textwidth}
\vspace{0pt}
\centering
\footnotesize
\setlength{\tabcolsep}{2.5pt}
\begin{tabular*}{\linewidth}{@{\extracolsep{\fill}}lrrrr@{}}
\toprule
Framework & $n$ & Ours & StepFinder & Last \\
\midrule
AlfAgent & 311 & 27.01 & 0.00 & 0.00 \\
Debate & 445 & 85.17 & 55.51 & 45.84 \\
DyLAN & 152 & 81.58 & 32.89 & 0.00 \\
MacNet & 261 & 50.96 & 49.04 & 0.00 \\
Magentic-One & 811 & 62.52 & 53.27 & 55.98 \\
MathChat & 920 & 89.35 & 76.30 & 78.80 \\
MetaGPT & 81 & 12.35 & 65.43 & 0.00 \\
SmolAgents & 3276 & 30.80 & 22.83 & 52.11 \\
\midrule
Framework macro & -- & 54.97 & 44.41 & 29.09 \\
Pooled & 6257 & 49.03 & 37.72 & 49.38 \\
\bottomrule
\end{tabular*}
\caption{Exact-step accuracy (\%) on new agent frameworks.}
\label{tab:lofo-step}

\end{minipage}
\end{table*}

\end{document}